\documentclass[letterpaper]{article} % DO NOT CHANGE THIS
\usepackage{aaai25}  % DO NOT CHANGE THIS
\usepackage{times}  % DO NOT CHANGE THIS
\usepackage{helvet}  % DO NOT CHANGE THIS
\usepackage{courier}  % DO NOT CHANGE THIS
\usepackage[hyphens]{url}  % DO NOT CHANGE THIS
\usepackage{graphicx} % DO NOT CHANGE THIS
\usepackage{natbib}  % DO NOT CHANGE THIS AND DO NOT ADD ANY OPTIONS TO IT
\usepackage{caption} % DO NOT CHANGE THIS AND DO NOT ADD ANY OPTIONS TO IT
\usepackage{algorithm}
\usepackage{algorithmic}
\usepackage{amsmath}
\usepackage{subfigure}
\usepackage{pifont}
\usepackage{multirow}
\usepackage{amssymb}

\usepackage{newfloat}
\usepackage{listings}
\DeclareCaptionStyle{ruled}{labelfont=normalfont,labelsep=colon,strut=off} % DO NOT CHANGE THIS
\floatstyle{ruled}
\newfloat{listing}{tb}{lst}{}
\floatname{listing}{Listing}
\title{PhysAug: A Physical-guided and Frequency-based Data Augmentation for Single-Domain Generalized Object Detection}
\author{
    Xiaoran Xu\textsuperscript{\rm 1,\rm 2},
    Jiangang Yang\textsuperscript{\rm 2}\equalcontrib,
    Wenhui Shi\textsuperscript{\rm 2},
    Siyuan Ding\textsuperscript{\rm 2},
    Luqing Luo\textsuperscript{\rm 2},
    Jian Liu\textsuperscript{\rm 1,\rm 2}\equalcontrib
}
\affiliations{
    \textsuperscript{\rm 1}School of Advanced Interdisciplinary Sciences, University of Chinese Academy of Sciences\\
    \textsuperscript{\rm 2}IPHAC Lab, Institute of Microelectronics of the Chinese Academy of Sciences\\
    xuxiaoran22@mails.ucas.ac.cn, 
    \{yangjiangang,luoluqing,shiwenhui,dingsiyuan,liujian\}@ime.ac.cn
}

\usepackage{bibentry}
\begin{document}

\maketitle

\begin{abstract}
Single-Domain Generalized Object Detection~(S-DGOD) aims to train on a single source domain for robust performance across a variety of unseen target domains by taking advantage of an object detector. Existing S-DGOD approaches often rely on data augmentation strategies, including a composition of visual transformations, to enhance the detector's generalization ability. However, the absence of real-world prior knowledge hinders data augmentation from contributing to the diversity of training data distributions. To address this issue, we propose PhysAug, a novel physical model-based non-ideal imaging condition data augmentation method, to enhance the adaptability of the S-DGOD tasks. Drawing upon the principles of atmospheric optics, we develop a universal perturbation model that serves as the foundation for our proposed PhysAug. Given that visual perturbations typically arise from the interaction of light with atmospheric particles, the image frequency spectrum is harnessed to simulate real-world variations during training. This approach fosters the detector to learn domain-invariant representations, thereby enhancing its ability to generalize across various settings. Without altering the network architecture or loss function, our approach significantly outperforms the state-of-the-art across various S-DGOD datasets. In particular, it achieves a substantial improvement of $7.3\%$ and $7.2\%$ over the baseline on DWD and Cityscape-C, highlighting its enhanced generalizability in real-world settings.
\end{abstract}

\begin{links}
    \link{Code}{https://github.com/startracker0/PhysAug}
\end{links}

\section{Introduction}

Object detection is a fundamental task in computer vision aiming to localize and recognize objects in natural images~\citep{ren2015faster,redmon2016you,tian2022fully,zou2023object}. In past decades, object detection has achieved significant progress due to the advancement of deep learning, which assumes that the training data and test data come from the same distribution~\citep{zhao2019object}. However, distribution shifts in various real-world applications make deep learning-based object detectors impractical. For instance, when trained on high-quality source data, object detectors for autonomous driving tend to perform poorly on different distributions of target data due to the changes in weather, illuminations, and object appearances. Such performance degradation hinders its wide deployment in safety-critical areas.

\begin{figure}[H]
    \centering
    \subfigure[]{
        \includegraphics[width=0.45\textwidth]{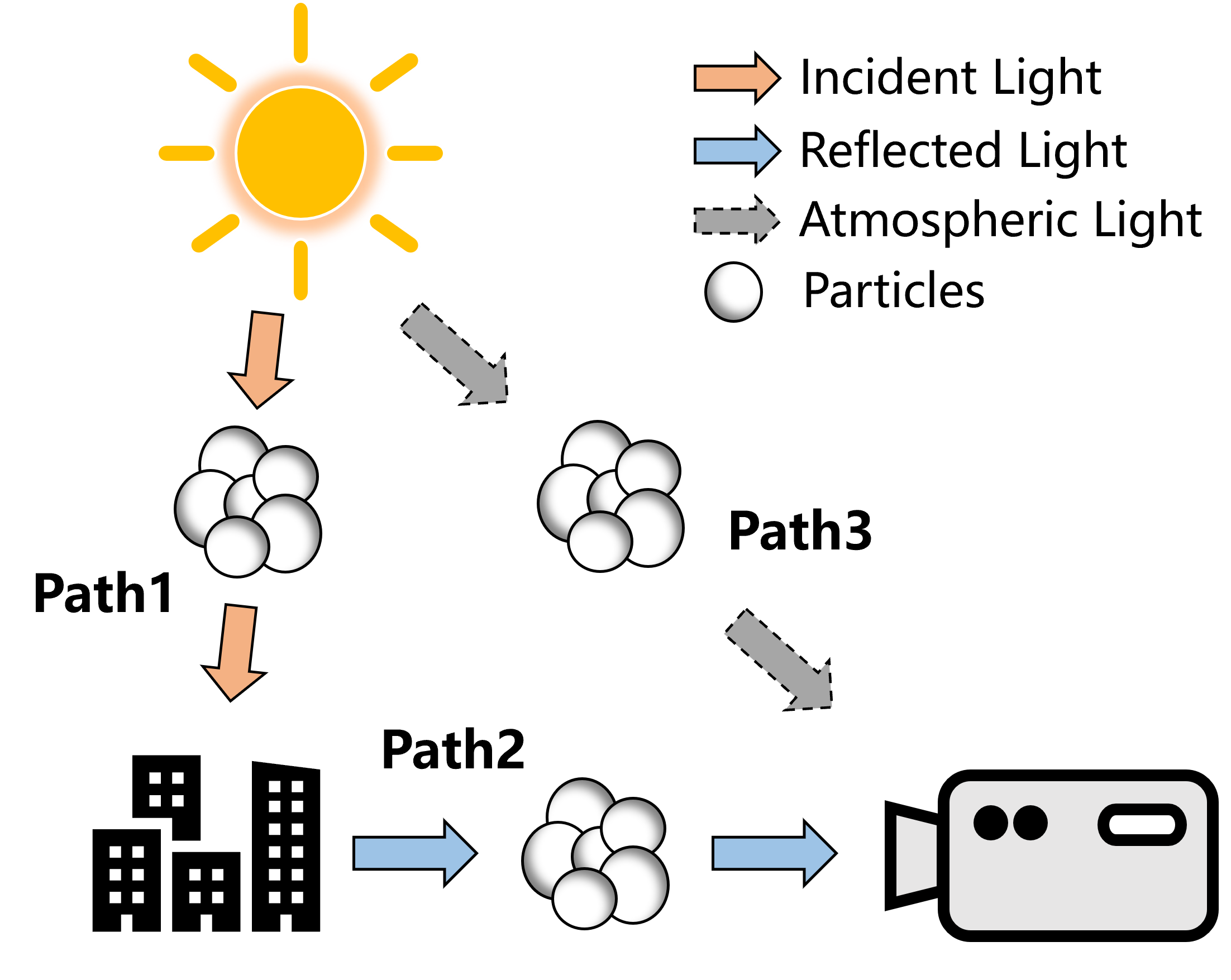}
        \label{fig1-a}
    }
    \hfill
    \subfigure[]{
        \includegraphics[width=0.45\textwidth]{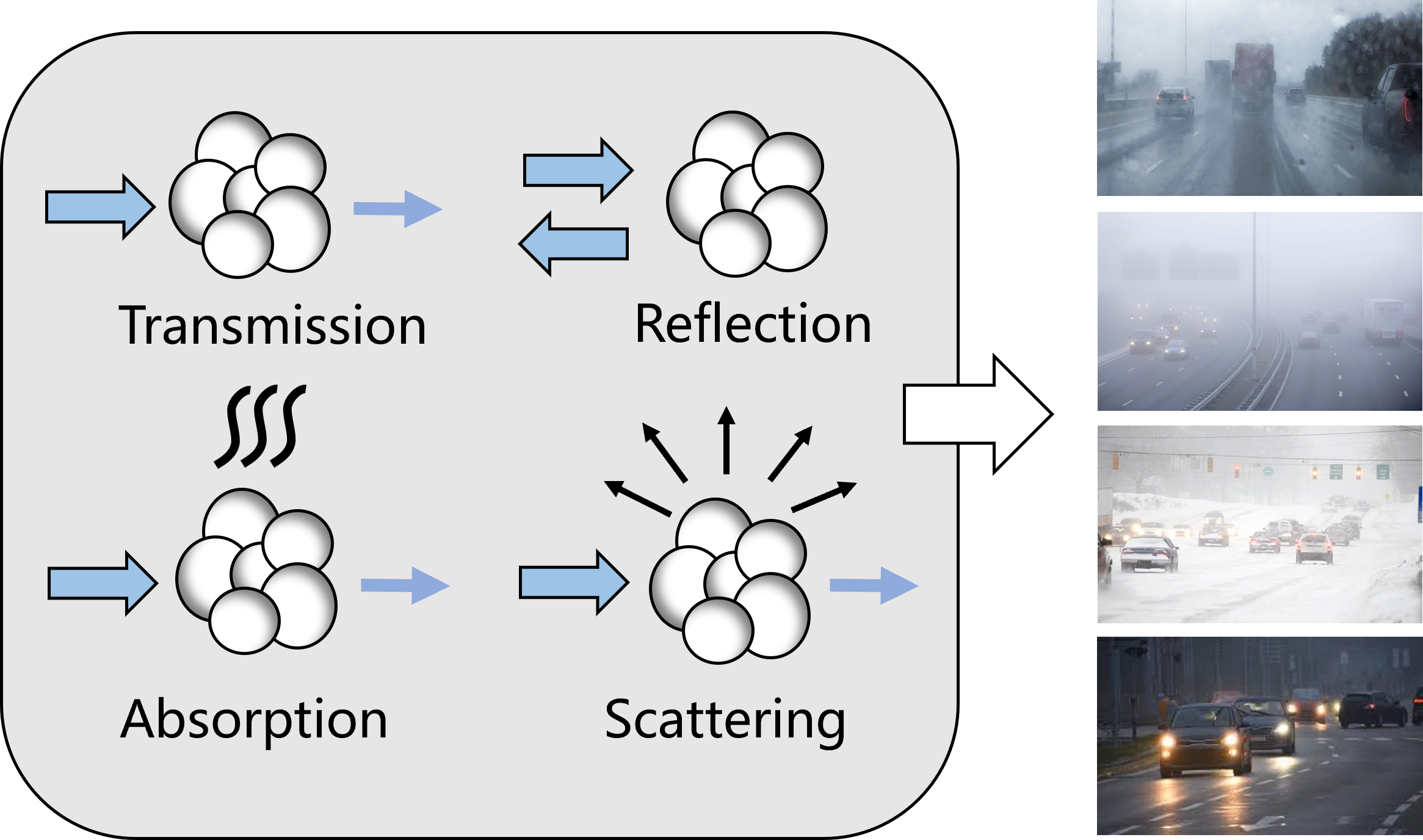}
        \label{fig1-b}
    }
     \caption{(a) The Physical imaging process in the atmosphere. Path1: Incident light path, Path2: Reflected light path, Path3: Atmospheric light path. (b) The cases of image degradation due to the different interactions between light and atmospheric particles.}
    \label{Figure 1}
\end{figure}

To tackle this issue, single-domain generalized object detection~(S-DGOD) has emerged as a prominent learning paradigm to alleviate the domain-shift impact, which aims at generalizing an object detector trained on a single source domain to multiple unseen target domains. Along this line of research, data augmentation is an active research direction for the S-DGOD problem to enrich the diversity of source domain. Several works propose to design the compositional policies of pre-defined visual corruptions applied on the global image or object-level regions~\citep{lee2024object} for expanding training samples. SRCD~\citep{rao2023srcd} introduces a mixing strategy of the global image with texture of its patches to enforce the model to focus on semantic content. UFR~\citep{liu2024unbiased} adapts a combination of frequency-based and spatial-based transformations to augment local objects. However, these methods do not explicitly consider the formation mechanism of non-ideal imaging conditions, preventing the capture of real-world variations. Fig.~\ref{fig1-a} depicts the optical imaging process with three light propagation paths through the atmosphere. Among these paths, as shown in Fig.~\ref{fig1-b}, interactions between light and various particles in the atmosphere are the primary causes of image degradation. 

In this paper, we first formulate a universal perturbation model from the theory of atmospheric optics. This perturbation model aims to explicitly characterize the inferior cases of optical imaging resulting from light propagation properties in the atmosphere. Guided by this physical-based model, we introduce our data augmentation method, namely PhysAug. Specifically, PhysAug is implemented from a fourier perspective: we adopt simple frequency-dependent filters to augment low-frequency components of the image, simulating the global non-uniform illumination caused by light scattering and attenuation. Furthermore, fourier basis functions are randomly combined and projected onto the spatial domain of the image, generating local occlusion effects due to non-transparent particles in the wild. We conduct experiments on two S-DGOD datasets, Diverse Weather Dataset~(DWD) and Cityscapes-C. The results show that PhysAug significantly outperforms existing state-of-the-art methods, achieving improvements of $7.3\%$ and $7.2\%$ over the baseline methods on the corrupted domains of DWD and Cityscapes-C, respectively. Besides, our analysis indicates that PhysAug enables models to learn better domain-invariant representations than non-physical based augmentations, proving its superiority. In summary, our contributions are as follows:
% include:
\begin{itemize}
\item We establish a universal perturbation model based on atmospheric optics, which is the first attempt to guide the design of data augmentation from the perspective of physical principles.

\item We introduce PhysAug, a frequency-based data augmentation method tailored for the S-DGOD problem. This method accurately simulates naturally-occurring visual degradation throughout the full trajectory of light propagation, enriching real-world variations in training data.

\item Comprehensive experiments and analysis validate the eﬀectiveness of our method. Our method achieves new state-of-the-art performance on two standard S-DGOD datasets, outperforming other approaches by a large margin.
\end{itemize}

\begin{figure*}[!ht]
    \centering
    \includegraphics[width=\linewidth]{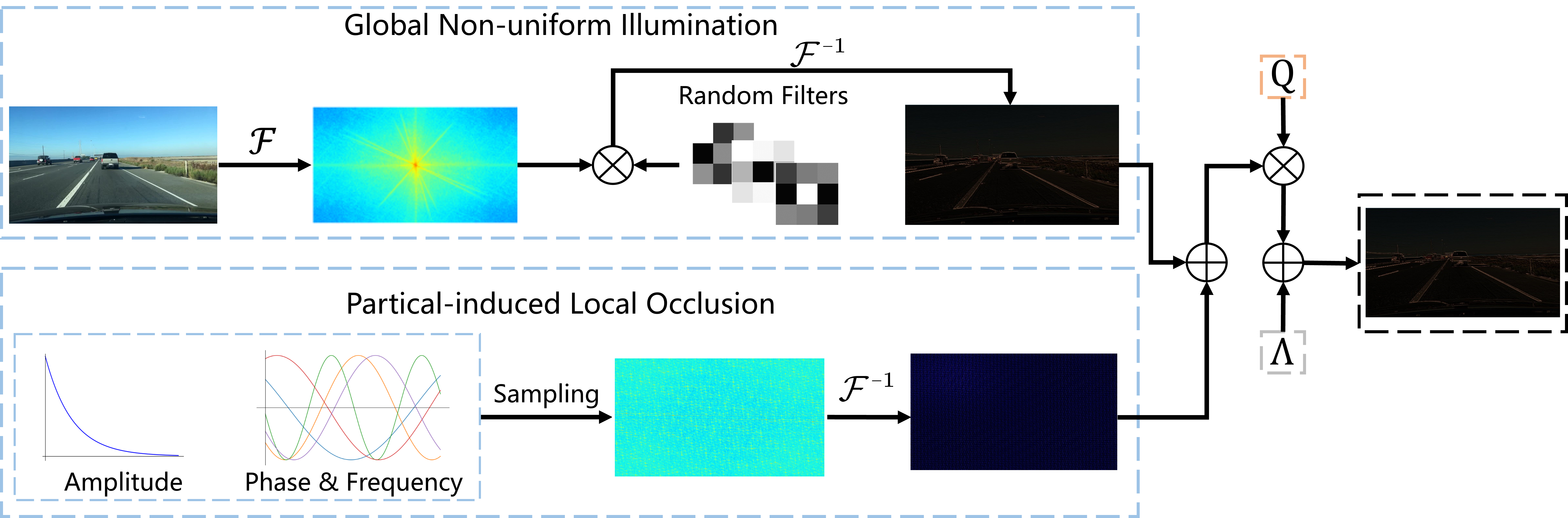}
    \caption{An overview of PhysAug. The input of PhysAug is an RGB training image, and its output is an augmented image. The two important components in PhysAug are Global Non-uniform Illumination~(Top) and Particle-induced Local Occlusion~(Bottom).}
    \label{Figure 2}
\end{figure*}

\section{Related Works}

\subsection{Single-Domain Generalization for Object Detection}
To enhance model performance in unseen target domains, recent advancements in this field have explored various approaches. These approaches could be categorized into three main types: data augmentation, losses for domain generalization, and network architecture improvements. 
Most existing methods address this task through data augmentation.  Those methods generate new domains via various transformations, thereby converting the S-DGOD problem into a domain adaptation object detection challenge. \citep{vidit2023clip} leveraged the vision-language model CLIP for semantic augmentation with textual prompts, enhancing the diversity of features extracted by the model. \citep{lee2024object} employs an object-aware augmentation approach that enables the model to better detect and focus on potential regions of interest. DivAlign~\citep{danish2024improving} similarly opts for diversity in the source domain and additionally aligns losses based on the class prediction results from the detector.
In addition to data augmentation, domain-invariant feature distillation and neural architecture search have also been successfully employed to enhance model generalization. CDSD~\citep{wu2022single} adopts a cyclic self-decoupling method to extract domain-invariant representations from the training data. G-NAS~\citep{wu2024g} searches for the most general architecture to prevent the model from overfitting. 
Unlike other methods, we propose PhysAug from the perspective of physical modeling, incorporating real-world prior knowledge into the augmentation process. 

\subsection{Data Augmentation for Domain Generalization}
The most commonly used method in domain generalization is data augmentation in image domain~\citep{zhou2022domain,shorten2019survey,zhao2020maximum,rebuffi2021data}, there are several widely applied methods, including MixUp~\citep{zhang2017mixup}, Cutout~\citep{devries2017improved}, and AugMix~\citep{hendrycks2019augmix}. Although image domain data augmentation has achieved significant success in enhancing model generalization and robustness. \citep{yin2019fourier} found that models trained with visual transformations could be susceptible to noise affecting specific parts of the frequency spectrum. AmpMix~\citep{xu2023fourier} linearly interpolates the amplitudes of two images while keeping the phase information unchanged to enhance the model's ability to generalize.
Furthermore, some studies attempt to combine image domain and frequency domain augmentation methods. PRIME~\citep{modas2022prime} employs a combined augmentation approach that integrates both frequency domain and image domain techniques. AFA~\citep{vaish2024fourier} builds upon the approach of PRIME and further expands the integration of image domain and frequency domain augmentation techniques.

\section{Preliminary}
\subsection{Problem Formulation}
Single-domain generalization for object detection~(S-DGOD) aims to train the model on a single source domain while ensuring it performs well across multiple target domains~\citep{wang2021robust}. Given a labeled source domain $D_s=\{\left( x_{i}^{s},y_{i}^{s} \right) \}_{i=1}^{N_s}$ and multiple unseen target domains $M=\left\{ D_{t}^{1},D_{t}^{2},\cdots ,D_{t}^{k} \right\}$, where $D_{t}^{n}=\{\left( x_{i}^{t} \right) \}_{i=1}^{N_t}$, $N_s$ and $N_t$ denote the size of the source and target datasets respectively. The S-DGOD task is to train an object detector $F$ on the source domain $D_s$, and test its performance across target domains $M$.

\subsection{Evaluation Metrics}
We follow the settings established by~\citep{michaelis2019benchmarking}, using mean average precision~(mAP) as metrics to evaluate the performance of our method on the clean domain. Furthermore, we also use mean performance under corruption~(mPC) established by~\citep{lee2024object} as metrics to evaluate our method's performance on corrupted domains, which represents the average of mean average precision~(mAP) across all corrupted domains and severity levels, the calculation is as follows.
\begin{equation}
mPC=\frac{1}{N_C}\sum_{C=1}^{N_C}{(\frac{1}{N_S}\sum_{S=1}^{N_S}{P_{C,S}})},
\label{Eq.(1)}
\end{equation}
$P_{C,S}$ is the target domain corrupted by corruption $C$ at severity $S$,  $N_{S}$ and $N_C$ means the number of severity and corruption. For DWD, the values for $N_s$, $N_c$ are $1$ and $4$, for Cityscapes-C, the values are $5$ and $15$.

\section{Methodology}
In this section, we introduce the design and implementation of PhysAug. First, we formulate the universal perturbation model from the theory of atmosphere optics. Based on the model, we derive our data augmentation method from a fourier perspective to explicitly simulate real-world variations for augmented data.

\subsection{A Physical-based Perturbation Model}
Physical imaging models based on atmospheric optics have achieved great success in various low-level computer vision tasks~\citep{li2019heavy}. Retinex theory~\citep{land1971lightness} is introduced to simplify the optical imaging process as a combination of the incident light image $L(x)$ and the reflected light image $R(x)$, concentrating on the light propagation of Path1 and Path2 in Fig.~\ref{fig1-a}. This can be represented by:
\begin{equation}
I\left( x,y \right) =L\left( x,y \right) \cdot R\left( x,y \right),
\label{Eq.(2)}
\end{equation}
where $I(x)$ is the observed image, $(x,y)$ is the image point. Retinex-based algorithms are designed to eliminate the effects of incident light to restore image quality, making them widely used for low-light enhancement tasks. The atmospheric scattering model~\citep{nayar1999vision} is another popular approach to model physical imaging in foggy conditions. Its mathematical form is as follows:
%--------------------------------
\begin{equation}
I\left( x,y \right) =t\left( x,y \right) \cdot J\left( x,y \right) +A\cdot \left[ 1-t\left( x,y \right) \right],
\label{Eq.(3)}
\end{equation}
%-------------------------------
where $J(x,y)$ represents the haze-free image, $A$ denotes the atmospheric light and $t(x,y)$ is the transmission function. The first term of Eq.~(\ref{Eq.(3)}) indicates light scattering effects along the reflected light path, corresponding to Path2 in Fig.~\ref{fig1-a}. The second term refers to changes in atmospheric light path~(e.g., Path3). The above modeling methods focus on special cases of the physical imaging process for image restoration. In practice, the quality of optical imaging is influenced by the interactions of light with different types of particles along all three propagation paths. So we formulate a perturbation model for covering general cases of inferior imaging, which is defined as:
\begin{align}
    &I\left( x,y \right) =Q\cdot \left[ h_g\left( x,y\right) +h_o\left( x,y\right) \right] +A\cdot t_w\left( x,y \right), \label{Eq.(4)} \\
    &h_g\left( x,y \right) =t_g\left( x,y \right) \cdot J\left( x,y \right), \label{Eq.(5)}\\
    &h_o\left( x,y \right) =t_o\left( x,y \right) \cdot \sum_{i=0}^n{P_i}, \label{Eq.(6)}
\end{align}
Let $Q$ denotes the incident light, and $A$ is the atmospheric light. $P_i$ represents the set of non-transparent particles, where $i$ indicates the particle type~(e.g., dust and water droplets) with $i \in [0,1,\cdots ,n]$. The transmission functions $t_g(x,y)$, $t_o(x,y)$ and $t_w(x,y)$ correspond to the reflected image $J$, the particle set $P_i$, and atmospheric light $A$, respectively. $ h_g(x,y)$ represents the global non-uniform illumination on the image $J(x,y)$ due to light scattering and attenuation. $h_o(x,y)$ reflects local occlusion stemming from light reflection and absorption by non-transparent particles. These two components together account for the potential perturbations along the reflected light path. This model satisfies two key properties: (1) Completeness. All three light propagation paths are covered. (2) Heterogeneity. It considers interactions between light and a vareity of particles. We use this model to guide the design of PhysAug.

\subsection{The Design of PhysAug}
\subsubsection{Overview}
In this section, we introduce the implemental details of PhysAug. Based on Eq.~(\ref{Eq.(4)}), we define PhysAug as follows:
\begin{equation}
\text{PhysAug}\left( x \right): =Q\cdot \left( \hat{h}_g\left( x \right) +\hat{h}_o\left( x \right) \right) +\varLambda,
\label{Eq.(7)}
\end{equation}
where $\hat{h}_g(\cdot)$ and $\hat{h}_o(\cdot)$ denote the estimates of $ h_g(\cdot)$ and $ h_o(\cdot)$, respectively. To simplify notation, we use $\varLambda $ to replace $A\cdot t_w\left( x,y \right) $. Fig.~\ref{Figure 2} depicts the overall structure of PhysAug consisting of four components: The most significant components are non-uniform illumination $ h_g$ and local occlusions $ h_o$ along the reflected light path, as most visual degradation generates in this path. We adopt frequency-based pipelines to simulate these two components, as shown in  Fig.~\ref{Figure 2}. Besides, $Q$ and  $\varLambda $ correspond to illumination changes from the incident light and atmospheric light path.
\subsubsection{Global Non-uniform Illumination}
Recent studies have shown that illumination information exhibits excellent separation properties in the image frequency domain, and is mainly concentrated in the low-frequency component~\citep{vaish2024fourier}. So we apply the convolution with random kernels to the image, adjusting the amplitude spectrum of the low-frequency components to generate a diverse range of non-uniform illumination conditions. This operation is defined as:
\begin{equation}
\begin{split}
\hat{h}_g\left( x,y \right) & := J\left( x,y \right) * G \\
& := \mathcal{F}^{-1}\left\{\mathcal{F}\left\{ J\left( x,y \right) \right\} \cdot \mathcal{F}\left\{G \right\}\right\}.
\end{split}
\label{Eq.(8)}
\end{equation}
where $\mathcal{F}$ and $\mathcal{F}^{-1}$ represent the Fourier transform and the inverse Fourier transform, respectively. $G$ denotes a convolutional filter with a kernel size of $n
\times n$, and its parameters are drawn from a gaussian distribution. $*$ is the convolution operation.
\subsubsection{Partical-induced Local Occlusion}
It is well known that visible light consists of components of different wavelengths. Here, we consider that local occlusions are generated from a composite of single-wavelength light which undergoes reflection and absorption by interacting with stmospheric particles. Inspired by previous fourier-based analysis~\citep{vaish2024fourier,chen2021amplitude,xu2023fourier}, we utilize the sinusoidal planar wave functions to simulate the occlusion case of single-wavelength light. The real part of planar wave function is defined as:
\begin{align}
    S^P\left( x,y \right) &=\Re \left( C\cdot e^{j2\pi f\cdot T(\omega,\phi)} \right),\label{Eq.(9)} \\
    T(\omega,\phi) &=  x\cos \left( \omega \right) +y\sin \left( \omega \right) +\phi,\label{Eq.(10)}
\end{align}
where $C$ is a constraint to enforce a unit l2-norm of the planar wave. $\Re(\cdot)$ denotes the real part. $T(\cdot)$ is a regular function. $f$ and $\omega$ represent the frequency and phase, respectively. $\phi$ is set to a constant of $\frac{\pi}{n}$, where $n\in \mathbb{N}$. $P$ denotes light of the specific wavelength, and in practice, we set $P\in [R, G, B]$. Hence, we use $S(x,y)=[S^{R}(x,y), S^{G}(x,y), S^{B}(x,y)]$ to represent the operations on R,G,B channels. Finally, we introduce the composite function to construct local occlusions from multiple types of particles:
\begin{equation}
\hat{h}_o\left( x,y \right) =\sum_{i=0}^n{M_i\cdot S_i\left( x,y \right)},
\label{Eq.(11)}
\end{equation}
where $i$ is the particle type. $M_i$ denotes a random matrix of the same size as the image for diversifying the inconsistent properties of local occlusions. According to Eq.~(\ref{Eq.(7)}), we implement PhysAug to increase the diversity of augmented data. Its parameter settings are provided in the implementation details. 

\begin{table*}[t!]
\centering
\fontsize{10}{12}\selectfont
% \vbox{}
\begin{tabular}{c|c|cccc|c}
\hline
        & Daytime Sunny & Night Sunny & Dusk Rainy & Night Rainy & Daytime Foggy & mPC \\ \hline
Baseline  & 50.4 & 37.5 & 29.2 & 14.6 & 33.1 & 30.2 \\
IBN-Net  & 49.7 & 32.1 & 26.1 & 14.3 & 29.6 & 25.5 \\
SW & 50.6 & 33.4 & 26.3 & 13.7 & 30.8 & 26.1 \\
IterNorm & 43.9 & 29.6 & 22.8 & 12.6 & 28.4 & 23.4 \\
ISW & 51.3 & 33.2 & 25.9 & 14.1 & 31.8 & 26.3 \\
SHADE & -- & 33.9 & 29.5 & 16.8 & 33.4 & 28.4 \\
CDSD & 56.1 & 36.6 & 28.2 & 16.6 & 33.5 & 28.7 \\
SRCD & -- & 36.7 & 28.8 & 17.0 & 35.9 & 29.6 \\
Vidit et al. & 51.3 & 36.9 & 32.3 & 18.7 & 38.5 & 31.6 \\
OA-Mix & 56.4 & 38.6 & 33.8 & 14.8 & 38.1 & 31.3 \\
OA-DG & 55.8 & 38.0 & 33.9 & 16.8 & 38.3 & 31.8 \\
Div & 50.6 & 39.4 & 37.0 & 22.0 & 35.6 & 33.5 \\
DivAlign & 52.8 &\underline{42.5} & \underline{38.1} & \textbf{24.1} & 37.2 & \underline{35.5} \\
UFR & \underline{58.6} & 40.8 & 33.2 & 19.2 & \underline{39.6} & 33.2 \\ \hline
PhysAug & \textbf{60.2} & \textbf{44.9} & \textbf{41.2} & \underline{23.1} & \textbf{40.8} & \textbf{37.5} \\ \hline
\end{tabular}

\caption{Comparison with state-of-the-art methods on the Diverse Weather Dataset~(DWD). OA-Mix and Div are data augmentation methods in OA-DG and DivAlign, respectively. The mPC indicates the average performance across four adverse weather conditions. Bold numbers represent the highest performance in each column, and underlined numbers indicate the second-highest rank. The results for Daytime Sunny are reported in mAP.}
\label{Table 1}
\end{table*}
 
\section{Experiments}
In this section, we first introduce the datasets and implement details. Then, we present our detailed experimental results on two standard S-DGOD datasets.
\subsection{Datasets}
We evaluate our approach on two large-scale object detection datasets, Diverse Weather Dataset~(DWD) and Cityscapes-C, following standard S-DGOD settings. DWD is an urban-scene detection dataset, which consists of five weather conditions, i.e., Daytime Sunny, Night Sunny, Night Rainy, Dusk Rainy and Daytime Foggy~\citep{wu2022single}. This Dataset is collected from multiple autonomous driving benchmarks~\citep{yu2020bdd100k,sakaridis2018semantic,hassaballah2020vehicle}. Daytime Sunny serves as a source domain containing 18,205 images for training and 8,313 for evaluation. Night Sunny, Night Rainy, Dusk Rainy and Daytime Foggy are used as unseen target domains, with $26,158$, $2,494$, $3,501$, and $3,775$ images, respectively. Cityscapes-C is a robust detection benchmark, which synthesizes $15$ types of common corruptions with five severity levels based on the validation set of Cityscapes~\citep{michaelis2019benchmarking}. These corruptions are categorized into four groups: Noise, Blur, Digital, and Weather. We use the training set of Cityscapes as the source domain, and corrupted images are regarded as unseen target domains for validation.

\subsection{Implementation Details}
We first describe the implementation details of the proposed PhysAug. We set the filter $G$ in Eq.~(\ref{Eq.(8)}) as the size of $3\times 3$ with a gaussian distribution $\mathcal{N}(0,4)$. The frequency $f$ and phase $\omega$ in Eq.~(\ref{Eq.(9)}) are both drawn from a uniform distribution $(-512, 512)$, and $\phi$ is $\frac{-\pi}{4}$. The matrix $M_i$ in Eq.~(\ref{Eq.(11)}) is sampled from 2D gaussian distribution $\mathcal{N}(0, \Sigma)$, where $\Sigma$ is an identical matrix with the same size as the image. $Q$ is set to $1$ to indicate a constant incident light coefficient. Following the atmospheric scattering model, $\Lambda$ represents the attenuation effects of atmospheric light, given by $\Lambda=L_{\infty}(1-e^{(-d)})$, and $L_{\infty}$ is the atmospheric light at infinity, which is set to $10^{-1}$~\citep{nayar1999vision}. We sample $d$ from a uniform distribution $(0,10)$. For DWD dataset, we adopt the setting up in CDSD~\citep{wu2022single} and OA-DG~\citep{lee2024object}. Specifically, we use Faster R-CNN~\citep{ren2015faster} with a ResNet101~\citep{he2016deep} backbone, and employ the SGD optimizer with a momentum of $0.9$ and a weight decay of $0.0001$. The learning rate is set to $0.001$, and the batch size is $2$. All other configurations align with those in the CDSD and OA-DG. For Cityscapes-C, we use Faster R-CNN~\citep{ren2015faster} with a ResNet50~\citep{he2016deep} backbone and feature pyramid networks~(FPN)~\citep{lin2017feature} as the baseline model. The learning rate is set to $0.01$, with a batch size of $8$. The SGD optimizer is employed with a momentum of $0.9$ and a weight decay of $0.0001$. All experiments are conducted on NVIDIA RTX 3090 GPU. 

\subsection{Performance on Diverse Weather Conditions}
We evaluate PhysAug and other the state-of-the-art methods on DWD dataset, including SW~\citep{pan2019switchable}, IBN-Net~\citep{pan2019switchable}, IterNorm~\citep{huang2019iterative}, ISW~\citep{choi2021robustnet}, SHADE~\citep{zhao2022style}, Clip the Gap~\citep{vidit2023clip}, OA-DG~\citep{lee2024object}, DivAlign~\citep{danish2024improving}, and UFR~\citep{liu2024unbiased}. Note that Div and OA-Mix are data augmentation methods in DivAlign and OA-DG, respectively. Table~\ref{Table 1} reports their performances on five different scenarios. We can see that PhysAug achieves the best detection accuracy of $60.2$~mAP in Daytime Sunny, $44.9$~mAP in Night Sunny, $41.2$~mAP in Dusk Rainy, and $40.8$~mAP in Daytime Foggy, respectively. Overall, our method obtains the largest improvement of $7.3$~mPC compared with the baseline detector in adverse weather conditions.

\begin{table*}[t]
\centering
\fontsize{9}{12}\selectfont % Set font size to 10pt with 12pt line spacing
\setlength{\tabcolsep}{0.8mm} % Adjust column width to fit within the page
\begin{tabular}{c|c|ccccccccccccccc|c}
\hline
\multirow{2}{*}{Method} & \multirow{2}{*}{Clean} & \multicolumn{3}{c}{Noise} & \multicolumn{4}{c}{Blur} & \multicolumn{4}{c}{Weather} & \multicolumn{4}{c|}{Digital} & \multirow{2}{*}{mPC} \\ \cline{3-17}
& & Gauss & Shot & Impulse & Defocus & Glass & Motion & Zoom & Snow & Frost & Fog & Bright & Contrast & Elastic & Pixel & JPEG \\ \hline
baseline & 42.2 & 0.5 & 1.1 & 1.1 & 17.2 & 16.5 & 18.3 & 2.1 & 2.2 & 12.3 & 29.8 & 32.0 & 24.1 & 40.1 & 18.7 & 15.1 & 15.4 \\ \hline
\multicolumn{18}{c}{\textbf{+Data augmentation}} \\ \hline
Cutout & 42.5 & 0.6 & 1.2 & 1.2 & 17.8 & 15.9 & 18.9 & 2.0 & 2.5 & 13.6 & 29.8 & 32.3 & 24.6 & \underline{40.1} & 18.9 & 15.6 & 15.7 \\
PhotoDistort & 42.7 & 1.6 & 2.7 & 1.9 & 17.9 & 14.1 & 18.7 & 2.0 & 2.4 & 16.5 & 36.0 & 39.1 & 27.1 & 39.7 & 18.0 & 16.4 & 16.9 \\
AutoAug & 42.4 & 0.9 & 1.6 & 0.9 & 16.8 & 14.4 & 18.9 & 2.0 & 1.9 & 16.0 & 32.9 & 35.2 & 26.3 & 39.4 & 17.9 & 11.6 & 15.8 \\
AugMix & 39.5 & 5.0 & 6.8 & 5.1 & 18.3 & 18.1 & 19.3 & \textbf{6.2} & 5.0 & 20.5 & 31.2 & 33.7 & 25.6 & 37.4 & 20.3 & 19.6 & 18.1 \\
StylizedAug & 36.3 & 4.8 & 6.8 & 4.3 & 19.5 & 18.7 & 18.5 & 2.7 & 3.5 & 17.0 & 30.5 & 31.9 & 22.7 & 33.9 & 22.6 & \underline{20.8} & 17.2 \\
OA-Mix & 42.7 & 7.2 & 9.6 & 7.7 & 22.8 & 18.8 & 21.9 & \underline{5.4} & 5.2 & \underline{23.6} & \underline{37.3} & 38.7 & \underline{31.9} & \textbf{40.2} & 22.2 & 20.2 & 20.8 \\ \hline
\multicolumn{18}{c}{\textbf{+Loss function}} \\ \hline
SupCon & \underline{43.2} & 7.0 & 9.5 & 7.4 & 22.6 & 20.2 & \underline{22.3} & 4.3 & 5.3 & 23.0 & \underline{37.3} & 38.9 & 31.6 & \underline{40.1} & 24.0 & 20.1 & 20.9 \\
FSCE & 43.1 & 7.4 & 10.2 & 8.2 & 23.3 & 20.3 & 21.5 & 4.8 & 5.6 & \underline{23.6} & 37.1 & 38.0 & \underline{31.9} & 40.0 & 23.2 & 20.4 & 21.0 \\
OA-DG & \textbf{43.4} & \underline{8.2} & \underline{10.6} & \underline{8.4} & \underline{24.6} & \underline{20.5} & \underline{22.3} & 4.8 & \underline{6.1} & \textbf{25.0} & \textbf{38.4} & \textbf{39.7} & \textbf{32.8} & \textbf{40.2} & \underline{23.8} & \textbf{22.0} & \underline{21.8} \\ \hline
\multicolumn{18}{c}{\textbf{Ours}} \\ \hline
PhysAug & 42.6 & \textbf{14.3} & \textbf{17.0} & \textbf{11.9} & \textbf{25.6} & \underline{19.1} & \textbf{25.5} & 3.9 & \textbf{8.6} & 21.3 & 35.3 & \underline{39.5} & 27.5 & 39.1 & \textbf{28.9} & 19.9 & \textbf{22.6} \\ \hline
\end{tabular}
\caption{The performance comparison with state-of-the-art methods on Cityscapes-C. The mPC is the average performance across $15$ corruption types, each evaluated at $5$ different severity levels. Numbers in bold represent the highest rank, and underlined numbers indicate the second highest rank.}
\label{Table 2}
\end{table*}

\subsection{Performance on Common Corruptions}
To further prove the generality of our proposed PhysAug, we also compare PhysAug with popular data augmentation approaches on Cityscapes-C, including Cutout~\citep{devries2017improved}, PhotoDistort~\citep{redmon2018yolov3}, AutoAug~\citep{zoph2020learning}, AugMix~\citep{hendrycks2019augmix}, StylizedAug~\citep{geirhos2018imagenet}. Besides, we select other three methods that combine data augmentation with loss function, SupCon~\citep{khosla2020supervised}, FSCE~\citep{sun2021fsce} and OA-DG~\citep{lee2024object} for comparison. These results are summarized in Table~\ref{Table 2}. It is noted that our proposed PhysAug achieves the state-of-the-art performance, improving by an average of $7.2$~mPC over the baseline detector. Besides, we can find that our method performs better on Noise and Blur corruption types, with the best result of $17.0$~mAP on Shot Noise, and $25.5$~mAP on Motion Blur, respectively.

\begin{table}[t]
    \centering
    \fontsize{9}{12}\selectfont
    \begin{tabular}{c|c|c|c}
        \hline
       Non-uniform Illumination & Local Occlusion & mAP & mPC \\
        \hline
        \multicolumn{4}{c}{DWD} \\
        \hline
         \ding{55} & \ding{55} & 50.4& 30.2\\
         \ding{51} & \ding{55} & 60.4 & 36.5 \\
         \ding{55} & \ding{51} & \textbf{60.6} & 35.2 \\
         \ding{51} & \ding{51} & 60.2& \textbf{37.5}\\
        \hline
        \multicolumn{4}{c}{Cityscapes-C} \\
        \hline
         \ding{55} & \ding{55} & 42.2 & 15.4 \\
         \ding{51} & \ding{55} & 42.4 & 20.5 \\
         \ding{55} & \ding{51} & \textbf{43.0} & 20.2 \\
         \ding{51} & \ding{51} & 42.6& \textbf{22.6}\\
        \hline
    \end{tabular}
    \caption{The performance comparison of each component in PhysAug on the DWD and Cityscapes-C datasets.}
    \label{Table 3}
\end{table}
%--------------------------------------------
\begin{table}[t]
    \centering
    \fontsize{10}{12}\selectfont
    \begin{tabular}{c|c|c|c}
        \hline
       & Physical-based modeling& mAP & mPC \\
        \hline
        Baseline &- & 42.2 & 15.4 \\
        NPM1 &\ding{55}& 42.5 &22.2  \\
        NPM2 &\ding{55}& 42.2 & 19.4 \\
        PhysAug &\ding{51}& \textbf{42.6}& \textbf{22.6}\\
        \hline
    \end{tabular}
    \caption{The comparison of physical-based and non-physical-based modeling on the Cityscapes-C dataset.}
    \label{Table 4}
\end{table}
%--------------------------------------------
\begin{figure*}[t]
    \centering
    \includegraphics[width=\linewidth]{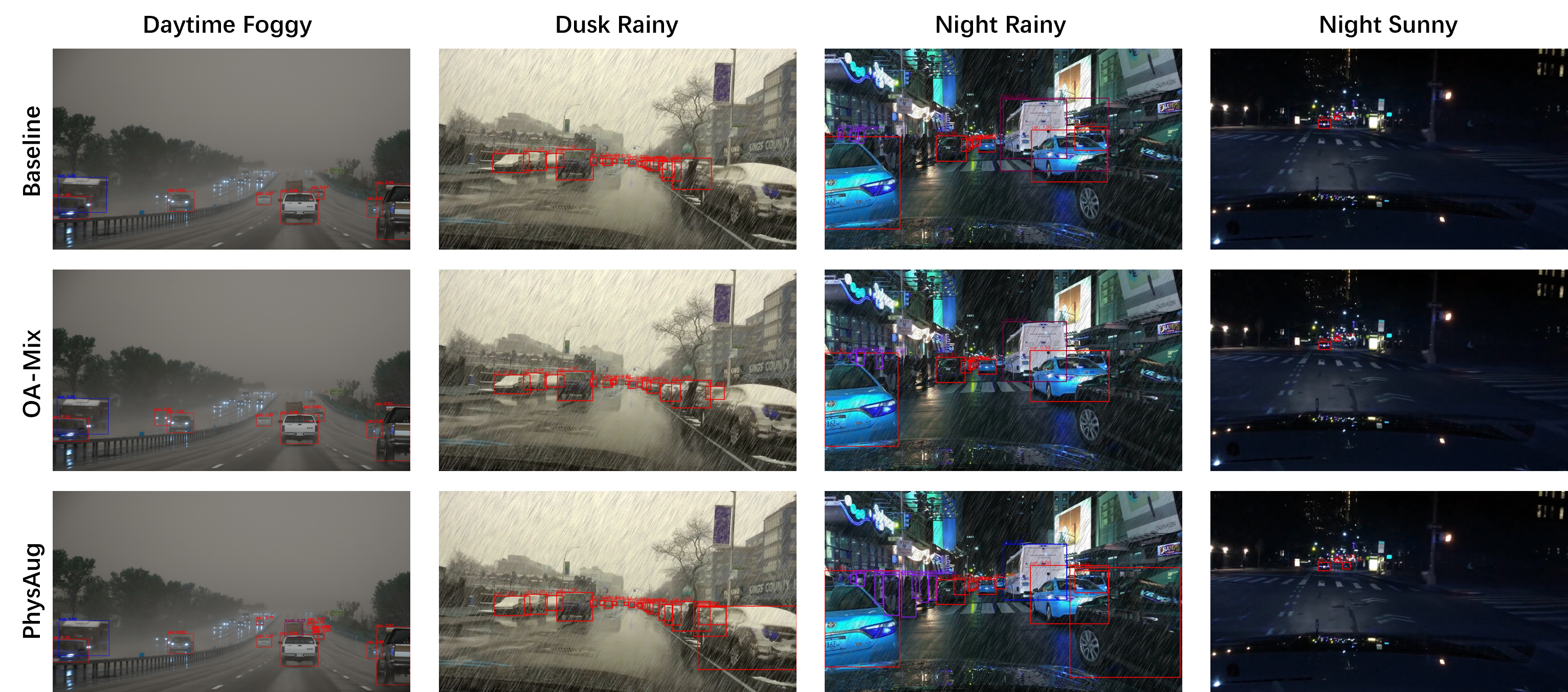}
    \caption{Visualized detection samples of the baseline method, OA-Mix and PhysAug in different weather conditions.}
    \label{Figure 3}
\end{figure*}
\begin{figure}[t]
    \centering
    \includegraphics[width=\linewidth]{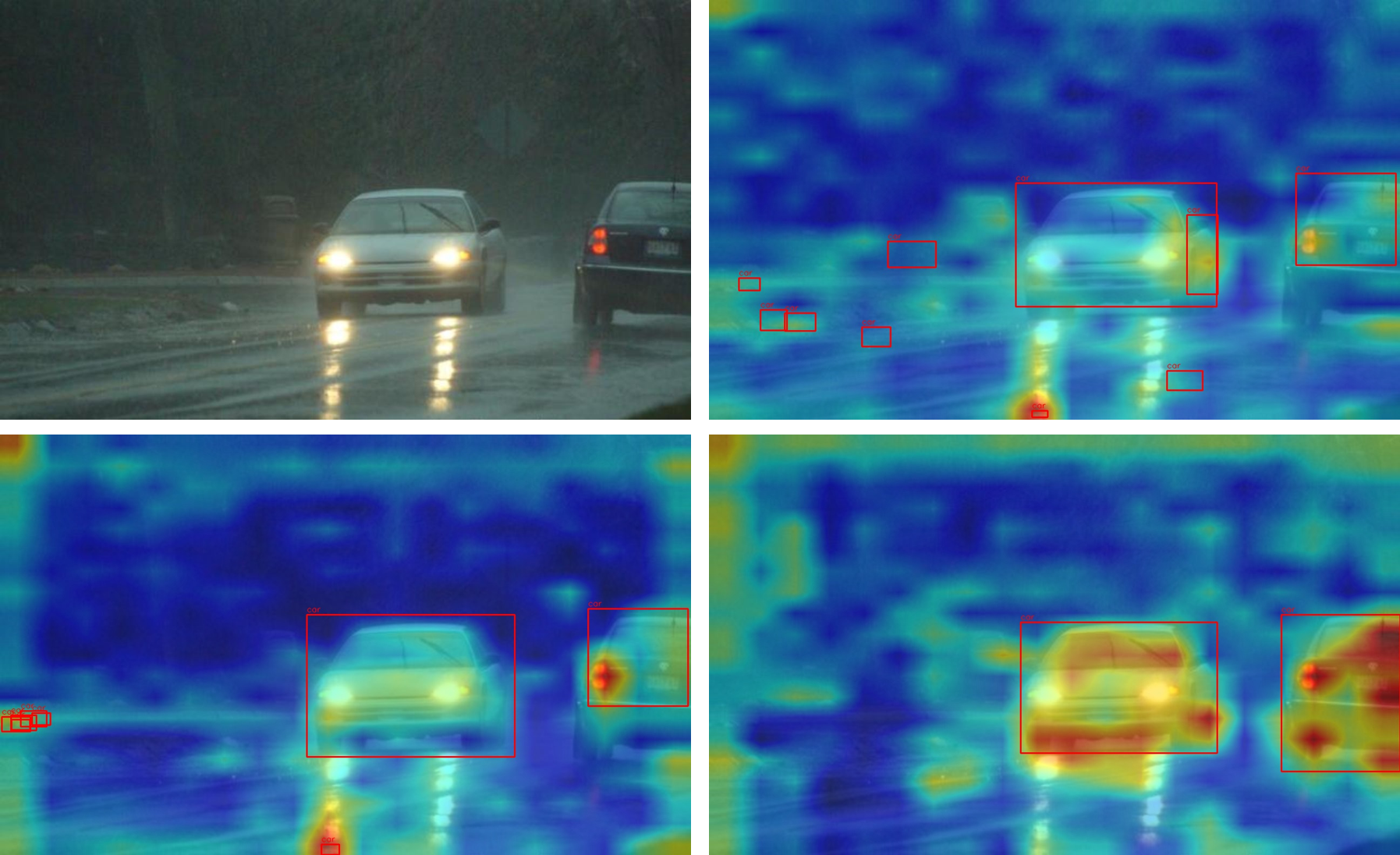}
    \caption{Heatmaps comparison in a night-rainy scenario. Top-left: original image. Top-right: The baseline method. Bottom-left: OA-Mix. Bottom-right: PhysAug.}
    \label{Figure 4}
\end{figure}

\section{Dicussion}
\subsection{Ablation Studies}
In this section, we conduct ablation studies to analyze the efficacy of the proposed PhysAug and highlight the benefits of guidance from physical-based modeling.

\subsubsection{PhysAug}
We now perform additional ablation experiments to study the impact of different components of the proposed PhysAug. Specifically, we compare the use of global non-uniform illumination in Eq.~(\ref{Eq.(8)}) and particle-induced local occlusions in Eq.~(\ref{Eq.(10)}). Table~\ref{Table 3} reports on our results on DWD and Cityscapes-C dataset. For DWD dataset, each component can still obtain improvements of above $5$~mPC in adverse weather conditions, indicating their versatility. The best performance is achieved with all components activated. One can see similar improvements of each component in Cityscape-C dataset, which validates the efficacy of the proposed method. 

\subsubsection{Physical-based Modeling}
This experiment aims to validate the benefits of physical-based modeling. So we construct two non-physical models for comparison. These models are as follows:
\begin{align}
    I\left( x,y \right) =& Q\cdot \left[ h_g\left( x,y \right) +h_o\left( x,y \right) \right],\label{Eq.(12)} \\
    I\left( x,y \right) =& \lambda\cdot Q\cdot  \left[ h_g\left( x,y \right) +h_o\left( x,y \right) \right] \nonumber\\
    &+(1-\lambda)\cdot J\left( x,y \right) +L\label{Eq.(13)}
\end{align}
where $\lambda$ is the mixing factor. we name the models derived from Eq.~(\ref{Eq.(12)}) and Eq.~(\ref{Eq.(13)}) as NPM1 and NPM2. NPM1 neglects the negative effect of the atmospheric light path, and NPM2 follows the MixUp-style~\cite{zhang2017mixup} to mix the non-corruption image based on Eq.~(\ref{Eq.(4)}). 
We compare the performance of PhysAug with NPM1 and NPM2 on the Cityscapes-C dataset, as shown in Table~\ref{Table 4}. The results show that PhysAug achieves the largest performance gains on corrupted domains, with $0.4-3.2$~mPC improvements on NPM1 and NPM2, respectively.

\subsection{Qualitative Analysis}
In this section, we further perform visualizations and feature assessment of our method to see how exactly it works.

\subsubsection{Visualization Analysis}
In Fig.~\ref{Figure 3}, we provide visualized object detection examples of the baseline method, OA-Mix and the proposed PhysAug on four adverse weather conditions. We can see that compared with the baseline method and OA-Mix, PhysAug can largely reduce the rate of false and missing detection in the image. Although night rainy weather is full of challenge, our method can localize and recognize these small and obscure persons accurately, which further demonstrates the effectiveness of the proposed PhysAug.

\subsubsection{Feature Analysis}
The heatmaps in Fig.~\ref{Figure 4} highlight the effectiveness of PhysAug in comparison to baseline and OAMix methods under challenging nighttime rainy conditions. The baseline method shows weak focus and sparse heatmap activation, indicating a limited ability to localize key objects such as vehicles in poor-visibility environments. OAMix improves target detection with more concentrated heatmap regions, particularly around the primary vehicle, but still leaves peripheral areas underrepresented. In contrast, PhysAug demonstrates significantly enhanced detection performance, with highly concentrated and precise heatmap activations around the vehicles under adverse visual conditions. These results suggest that PhysAug effectively captures spatial and frequency-based features, resulting in superior object localization and robustness.

\subsection{Conclusion}
In this paper, we propose PhysAug, a novel data augmentation method for single-domain generalized object detection. To achieve this goal, we introduce a universal perturbation model from the theory of atmospheric optics to describe the general case of the inferior imaging process. Guided by this model, PhysAug simulates real-world variations caused by atmospheric particles to diversify training data. We conduct extensive qualitative and quantitative experiments to demonstrate the effectiveness of our method for the S-DGOD task. Results also show that PhysAug outperforms the state-of-the-art by a large margin.

\section{Acknowledgements}
This work was supported by the National Key Research and Development Program of China under Grant 2021YFB2501403.

\bigskip

% \bibliography{aaai25}

\end{document}